\documentclass[conference]{IEEEtran}
\IEEEoverridecommandlockouts
\usepackage{cite}
\usepackage{amsmath,amssymb,amsfonts}
\usepackage{algorithmic}
\usepackage{graphicx}
\usepackage{hyperref}
\usepackage{float}
\usepackage{textcomp}
\usepackage{graphicx}
\usepackage{subcaption}
\usepackage{xcolor}
\def\BibTeX{{\rm B\kern-.05em{\sc i\kern-.025em b}\kern-.08em
    T\kern-.1667em\lower.7ex\hbox{E}\kern-.125emX}}

\begin{document}

\title{Enhancing Robustness of Indoor Robotic Navigation with Free-Space Segmentation Models Against Adversarial Attacks}

\makeatletter
\newcommand{\linebreakand}{%
  \end{@IEEEauthorhalign}
  \hfill\mbox{}\par
  \mbox{}\hfill\begin{@IEEEauthorhalign}
}
\makeatother
\author{\IEEEauthorblockN{Qiyuan An, Christos Sevastopoulos, Fillia Makedon
\linebreakand
}
\IEEEauthorblockA{Department of Computer Science \& Computer Engineering, University of Texas at Arlington, Arlington, TX, USA \ \\
\
\{qxa5560, christos.sevastopoulos\}@mavs.uta.edu}
makedon@uta.edu

}

\maketitle


\begin{abstract}
Endeavors in indoor robotic navigation rely on the accuracy of segmentation models to identify free space in RGB images. However, deep learning models are vulnerable to adversarial attacks, posing a significant challenge to their real-world deployment. In this study, we identify vulnerabilities within the hidden layers of neural networks and introduce a practical approach to reinforce traditional adversarial training. Our method incorporates a novel distance loss function, minimizing the gap between hidden layers in clean and adversarial images. Experiments demonstrate satisfactory performance in improving the model's robustness against adversarial perturbations.
\end{abstract}


\section{Introduction}

Free-space segmentation is a vital concept in robotics and autonomous driving, particularly in the context of path planning for mobile agents. This involves dividing the traversable space into distinct segments, that the autonomous agent can safely navigate \cite{sevastopoulos2022survey}. These segments are typically determined based on the physical layout of the environment, such as obstacles, terrain, and other features that might restrict motion~\cite{manduchi2005obstacle}. The goal of free-space segmentation is to enable the agent to plan an optimal path while avoiding obstacles and navigate through the environment safely and efficiently. 

Contemporary research leverages deep learning techniques, with substantial potential for this task. By employing pre-trained models and transfer learning, mobile agents learn to identify traversable regions using positive instances representing obstacle-free areas~\cite{hirose2019deep,b29}. Nevertheless, challenges such as the stochastic nature of the environment, presence of humans, variations in lighting conditions and occlusions can largely affect the perception of free space \cite{b34}. Furthermore, indoor environments pose unique difficulties, with their inclusion of ambiguous and semantically diverse objects such as furniture, household items, and interior structures. These elements may exhibit textures similar to traversable surfaces, potentially leading to inconsistent predictions. 
\begin{figure}[t]
\centering
\includegraphics[width=0.48\textwidth]{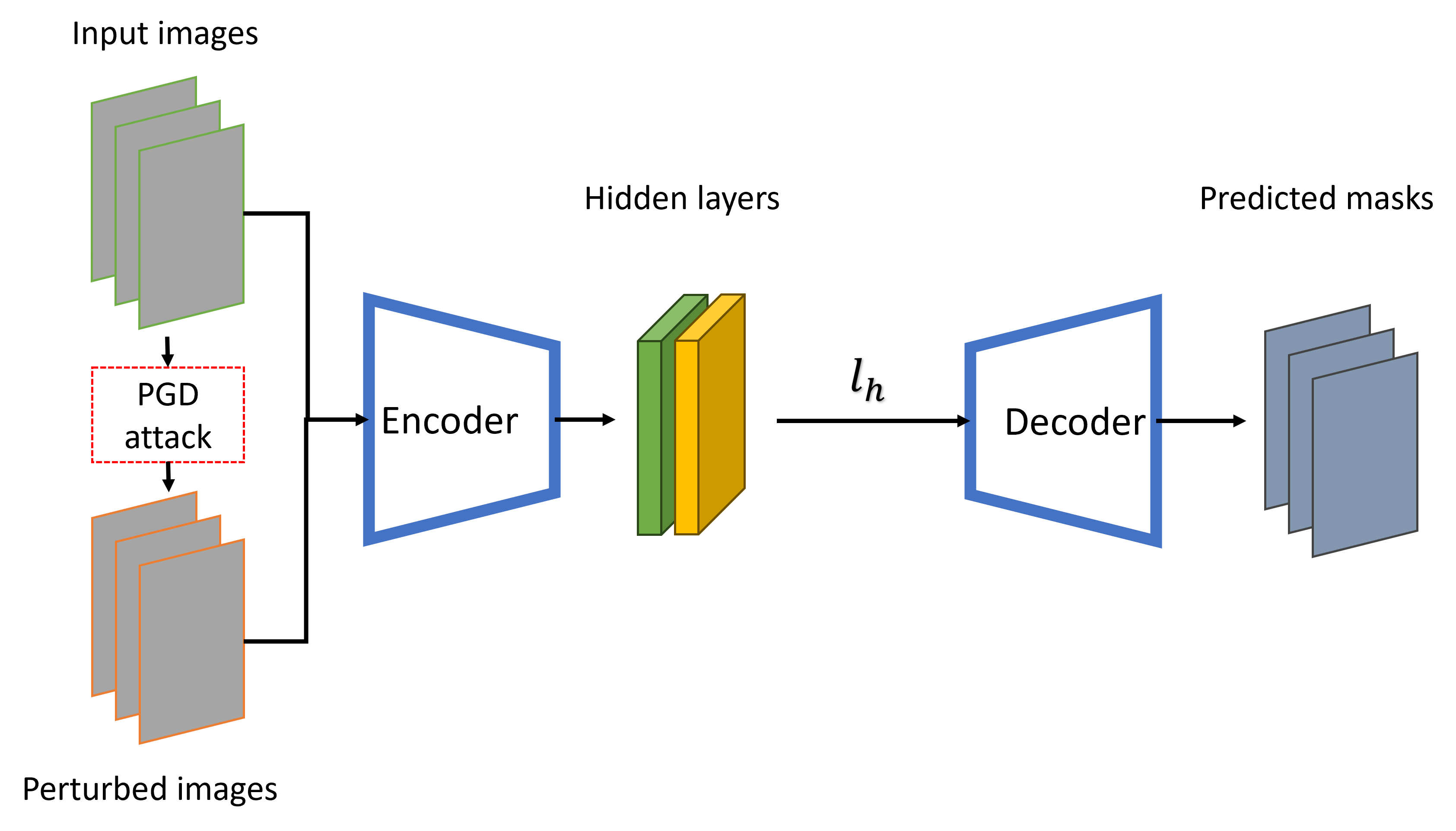}
\caption{Proposed Methodology}
\label{fig:arch}
\end{figure}

Adversarial attacks are a type of threat to the security of deep learning models~\cite{moosavi2016deepfool}. These attacks involve creating imperceptible perturbations to human eyes on the input but can greatly alter the model's output. The attacker leverages deep learning models' gradients to create gradients with respect to the input and produces incorrect outputs, with the goal of causing harm or danger consequently. In autonomous navigation, the result of an adversarial attack could cause an agent to have erroneous perception of the environment, and thus significantly jeopardizing its safety~\cite{deng2020analysis}.

In this article, the main goal is to tackle adversarial perturbation for an indoors free-space segmentation task. Our primary hypothesis relies on applying regularization techniques to intermediate network layers rather than exclusively applying supervisory signals on the network's final layers~\cite{wong2018provable}. Hence, we can effectively mitigate the gradual increasing effect of harmful adversarial perturbations. 
We achieve this by modifying the adversarial training loss function, i.e. introducing an additional term that minimizes the divergence in the embedding space between adversarial and clean images. 

To test the proposed method, we evaluate on our custom-collected indoor dataset. In this dataset, RGB and 1D laser measurements are collected using a mobile robotic platform and we employ an automated annotation tool that takes advantage of both velocity and laser readings. This approach allows to accurately label the frames of the encountered scenes according to the positive and unlabeled (PU) method~\cite{b49}, that learns from positive and unlabeled data. Finally, we test on challenging instances and report the method's performance.
 Code and dataset are available at \href{https://github.com/qiyuan-an/segadv.git}{https://github.com/qiyuan-an/segadv.git}

\section{Related work}

Extensive research has been conducted on predicting the geometry of scenes using RGB-D information, with a primary emphasis on outdoor scenarios~\cite{b22,b25}, yielding valuable insights into the identification of traversable free-space locations~\cite{b20,b24}. In indoor settings, notable efforts have been made to assess scene traversability at a higher conceptual level\cite{b21,b26}, yet without delving into the segmentation of free-space in detail.


Adversarial attacks can target the sensors used by autonomous agents, such as LIDARs~\cite{cao2019adversarial} and cameras~\cite{lu2017no}. Attackers introduce subtle perturbations to sensor data, leading to false interpretations by the perception systems~\cite{wang2019adversarial}. 
Mitigating the threat of adversarial attacks poses a substantial challenge in the field of machine learning. To address this issue, researchers have put forth several defense mechanisms. Adversarial training \cite{szegedy2013intriguing}, involves augmenting the training dataset with adversarial examples to enhance the overall model robustness. Another approach is \emph{input transformation}~\cite{guo2017countering}, where input data are modified to become more resilient to adversarial perturbations. Additionally, \emph{model distillation}~\cite{papernot2016distillation} is a technique that involves smoothing the model's output to reduce sensitivity to high-frequency perturbations. Achieving improved adversarial performance necessitates a significantly larger model capacity, as described in~\cite{xie2019intriguing}. 


Such attacks are primarily directed towards image classification, as the perturbations are often imperceptible yet can result in significant changes in the classification outcomes. Examples include the \emph{one pixel attack}, which can alter the classification label~\cite{su2019one}, the use of \emph{adversarial T-shirt patches} to deceive face recognition surveillance cameras~\cite{xu2020adversarial}, and the application of adversarial patches to mislead traffic sign detection systems in autonomous driving~\cite{liu2019perceptual}. 

Kannan et al. ~\cite{kannan2018adversarial} implement a mixed version of adversarial training~\cite{goodfellow2014explaining,madry2017towards} by incorporating both clean and perturbed images into batches rather than solely training on perturbed images. Nevertheless, their method degrades when the adversary strength increases in the image classification task as noted in~\cite{engstrom2018evaluating}. This last aspect is crucial for our task. We need to carefully select the strength of applied perturbation, while remaining imperceptible, in order to affect the performance of free-space segmentation task. Consequently, we only focus on negligible perturbations that can still largely affect the segmentation results.

\section{Methodology}

\subsection{Overview}

Overall, our objective is to design an adversarial loss function in a way that enhances the segmentation model's robustness.  The proposed architecture is illustrated in Figure~\ref{fig:arch}. We first train a SegFormer model \cite{xie2021segformer} on the positive subset of our dataset which consists of instances that represent obstacle-free areas. Afterwards, in the spirit of traditional adversarial training, adversarial examples are generated from the positive subset through performing a Projective Gradient Descent (PGD) attack \cite{madry2017towards}. Moreover, we introduce an adversarial hidden loss, that aims to minimize the divergence within the hidden layers of the SegFormer's encoder's output, between clean and adversarial images. 
This modification extends the pixel-wise cross-entropy loss function by incorporating an additional term that reduces the difference of hidden layers between clean and adversarial examples.
Finally, the SegFormer model is fine-tuned on mini-batches consisting of pairs of clean and adversarial examples.

\subsection{Traditional Adversarial Training}

At the core of our method is the application of traditional adversarial training~\cite{madry2017towards} that has shown substantial defensive efficiency against adversarial attacks. The rationale behind the  integration of adversarial examples into the training process is that it enables the model to improve its generalization capabilities and reduce its susceptibility to minor perturbations in input data~\cite{kannan2018adversarial}.
PGD attack generates adversarial examples, and then the model is trained with a mixture of both original and generated examples. While the downstream model learns features from both clean and adversarial inputs, it only partially aligns with the approach described in Eq.~\ref{eq:minimax}.
This is due to its inability to apply the supervisory signal solely from the end of the neural network to regulate the network's output. Therefore, our proposed approach aims to employ the supervisory signal at the intermediate network's layers, for the sake of moderating the disparities between the hidden layers of clean and adversarial examples.
\begin{equation}
L=\min_{\theta} \mathbf{E}_{(x,y)\sim\mathcal{D}}[\max_{\delta\in S}L(\theta, x+\delta, y)],
\label{eq:minimax}
\end{equation}

where $\theta$ denotes the downstream model's parameters, $x$ denotes clean images, $\delta$ represents generated perturbations, $y$ signifies the correct labels, and $L$ is the primary task loss.


\subsection{Incorporating an adversarial hidden loss}

We aim to maximize the mutual information between the hidden layers of clean and adversarial examples~\cite{oord2018infonce}. This approach enables us to extract the underlying shared information present in the inputs.
Accordingly, we introduce the adversarial hidden loss, denoted as $l_h$, as an enhancement to traditional adversarial training. This loss function is designed to reduce the distance between clean and adversarial examples at the hidden layer level (Eq.~\ref{eq:hidden_loss}).
\begin{equation}
l_h=\min_{\theta}L(\theta, x, x+\delta).
\label{eq:hidden_loss}
\end{equation}

Therefore, the total loss function combines the traditional adversarial training loss and our proposed adversarial hidden loss as follows:
\begin{equation}
l_{total}=L+\lambda l_h,
\label{eq:total_loss}
\end{equation}
where $\lambda$ is a hyperparameter to control the regularization strength.

\section{Experimental Setup}

\subsection{Data Collection}

For the data collection, a human operator directly teleoperated the Summit-XL Steel robotic platform with a wireless PS4 controller. We used ROS Melodic\footnote{\url{http://wiki.ros.org/melodic}}
and the \textit{message\_filters} data synchronization package\footnote{\url{http://wiki.ros.org/message filters}}.
to operate the robot and to record the
RGB, laser range finder, IMU, and encoder channels.

\subsection{Data Annotation}

The robot was deployed in various buildings on the University of Texas, Arlington (UTA) campus, where it navigated hallways and areas with changing lighting conditions and diverse objects. To ensure safety, the operator stopped the robot if there was a potential interference with humans or objects. In total, we collected a dataset of 3324 monocular RGB images.

Utilizing the Positive-Unlabeled (PU) method~\cite{b49}, we divide each source set (corresponding to a specific university building) into two subsets: positive and unlabeled image instances. However, relying solely on wheel velocity for automated annotation could lead to inaccuracies, especially in scenarios with unexpected human presence or limited odometry information. To enhance the reliability of annotation, we incorporated information from a laser scanner.

The annotation process involved examining the velocities of all four wheels, establishing a threshold value of 1 m/s, and a time window of 2.5 seconds. We defined three conditions: the wheel velocities consistently exceeding the threshold, the robot moving forwards or turning, and no obstacles detected within a 1.2-meter range of the laser sensor. If these conditions were met, we labeled the central frame within the time window as "positive" representing a traversable scene. Frames that did not meet the conditions were left unlabeled. From the unlabeled set, we manually removed scenes where the robot had limited or no free space to proceed, and labeled scenes with both free space and obstacles as "challenging". Thus, our dataset for experiments consisted of 2553 images, including 2010 positive and 443 challenging instances.

\begin{figure}[h]
\centering
\includegraphics[width=0.49\linewidth]{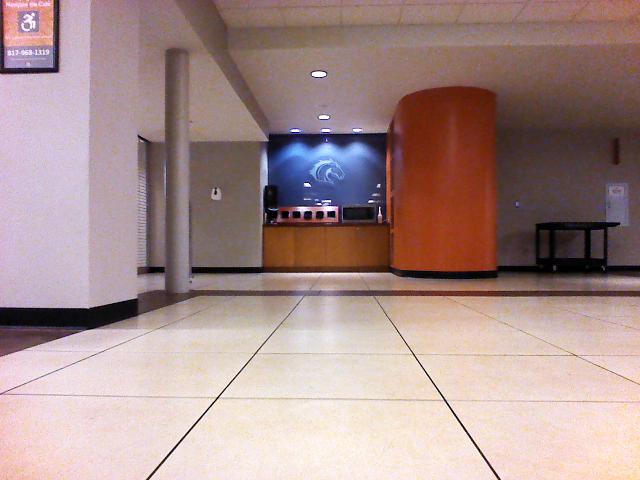}
\hfill
\includegraphics[width=0.49\linewidth]{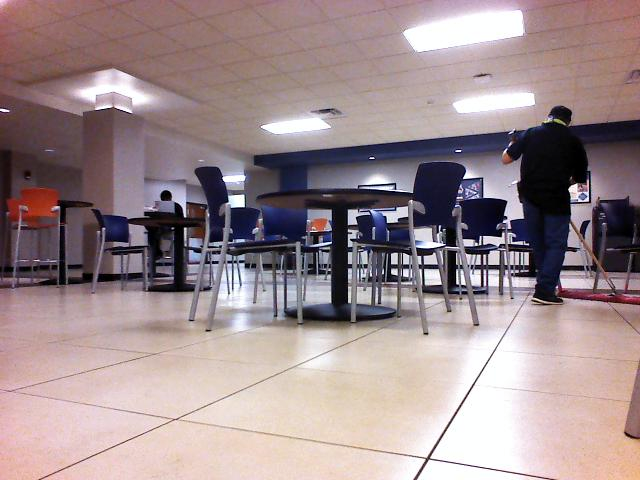}
\caption{Illustrative examples of positive and challenging instances. For the positive example (left) the robot is free to traverse while in the challenging case (right), the free space is limited due to the presence of static (chairs/tables) and dynamic (humans) objects. }
\label{fig:examples}
\end{figure}

\begin{figure}[H]
\centering
\begin{subfigure}[b]{0.22\textwidth}
\includegraphics[width=\textwidth]{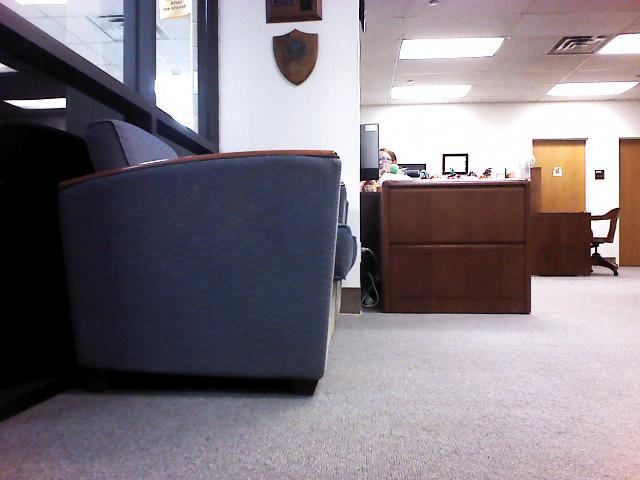}
\caption{$\epsilon$ = 0.005}
\end{subfigure}
\begin{subfigure}[b]{0.22\textwidth}
\includegraphics[width=\textwidth]{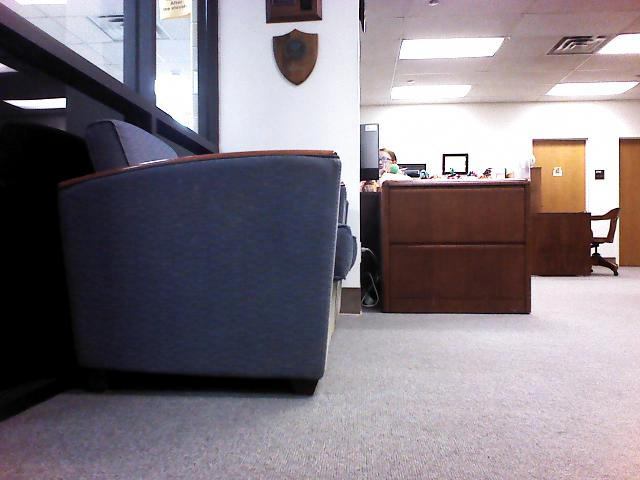}
\caption{$\epsilon$ = 0.01}
\end{subfigure}
\vfill
\begin{subfigure}[b]{0.22\textwidth}
\includegraphics[width=\textwidth]{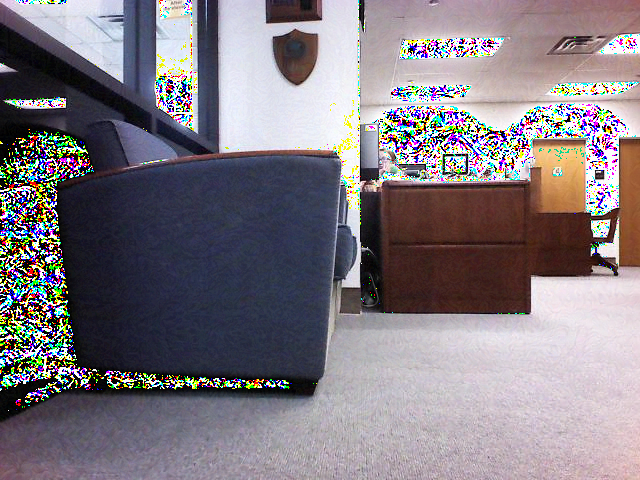}
\caption{$\epsilon$ = 0.05}
\label{fig:image3}
\end{subfigure}
\begin{subfigure}[b]{0.22\textwidth}
\includegraphics[width=\textwidth]{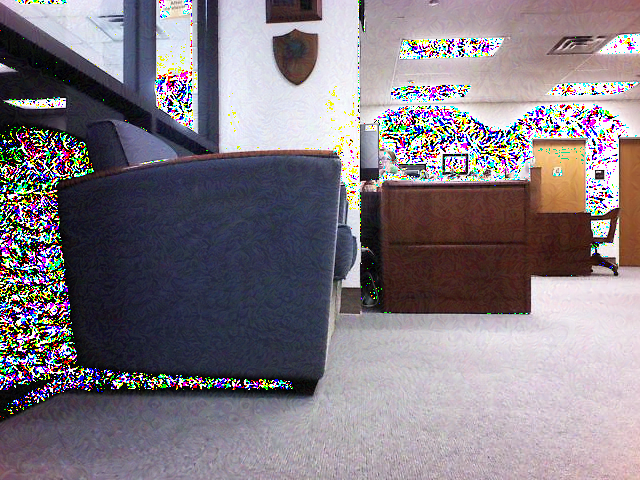}
\caption{$\epsilon$ = 0.1}
\label{fig:image4}
\end{subfigure}
\caption{An example of generated perturbations for different $\epsilon$ values}
\label{fig:epsilons}
\end{figure}

\subsection{Fine-tune a SegFormer}

We fine-tune a SegFormer model on our custom-collected dataset. The SegFormer is merging the Transformer's architectural backbone with a lightweight decoder. In contrast to ViT, its~\emph{Mix Transformer} encoder (MiT) does not use any positional encodings and can generate multi-level feature maps (both high-resolution fine features and low-resolution coarse features) due to its hierarchical structure. Furthermore, a series of lightweight (fewer number of parameters) Multi-Layer Perceptrons (MLP) is used as a decoder, which exhibits the attribute of combining both local and global attention, and eventually creates powerful and meaningful representations accompanied by strong generalization capabilities. This last aspect, along the dominant multi-scale feature learning ability of the SegFormer's encoder, can be proven to be crucial for our method; specifically, we use a SegFormer (MiT-B5) pre-trained on the Cityscapes dataset, and replace the classification head for fine-tuning on our dataset.

\begin{table*}[h]
\caption{Performance (mIoU) for different model configurations}
\centering
\begin{tabular}{c|c|c|c|c|c}
\hline
\textbf{PGD Attack} & \textbf{Loss Function} &\textbf{Adversarial Training}  & \textbf{SegFormer} & \textbf{DeepLabv3} & \textbf{PSPNet} \\
\hline
\hline

$\times$ & Pixel-wise cross-entropy & $\times$&  0.881 & 0.862 & 0.856 \\
\hline
\checkmark & Pixel-wise cross-entropy& $\times$ & 0.787 & 0.738 & 0.73 \\
\hline
\checkmark & Pixel-wise cross-entropy& \checkmark & 0.835 & 0.801 & 0.814 \\
\hline
\checkmark & Pixel-wise cross-entropy + Adversarial hidden &\checkmark&  0.865 & 0.836 & 0.828 \\ 
\hline
\end{tabular}
\label{tab:quan}
\end{table*}

\subsection{Implementation Details}

We used the PyTorch\footnote{\url{https://pytorch.org}} framework as the basis of our experiments. Training was done on a machine with 2 Titan RTX GPUs (24GB GDDR6 RAM, 4608 Cuda Cores). Using the pixel-wise cross-entropy loss function, we trained for 50 epochs unless an early stopping callback terminated the trial upon observed convergence. As training parameters we used: batch size = 16, learning rate = 0.01 and weight decay = 5e-4. The fine-tuning of the layers was accomplished using stochastic gradient descent (SGD). All images' initial dimensions of 640x480 pixels were re-scaled to 224x224 using the default PyTorch interpolation. In the following sections, we present the best results achieved when using 1628 of positive instances for training, and 492 challenging instances for testing.

\section{Results and discussion}




As the perturbation strength is intensified, the adversarial effect becomes more evident (Figure~\ref{fig:epsilons}). Therefore, our focus is solely placed on slight perturbations that can still significantly impact the segmentation results. Hyperparameter $\epsilon$ in the PGD attack, limits the magnitude of the perturbation that can be applied to the input data \cite{madry2017towards}. It is important to note that the threshold value is set at 0.01, since higher values can significantly alter the input image.


\begin{figure*}[p]
\centering
\begin{subfigure}{\textwidth} 
\centering
\begin{subfigure}{0.24\linewidth}
\centering
\subcaption[]{CASE 1}
 \includegraphics[width=\linewidth]{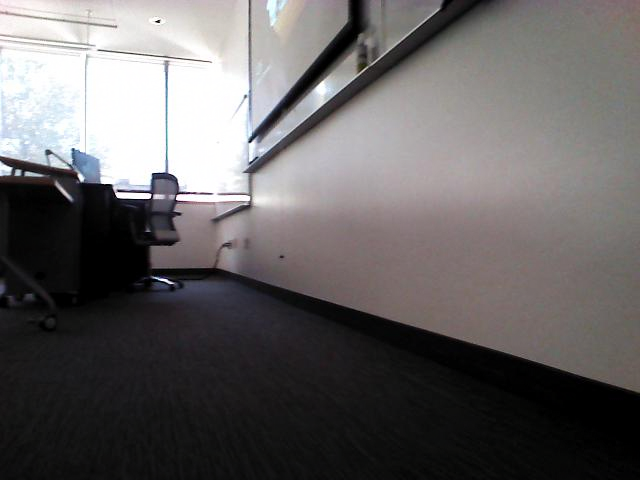}
\end{subfigure}
\hfill
\begin{subfigure}{0.24\linewidth}
\centering
\caption{CASE 2} 
\includegraphics[width=\linewidth]{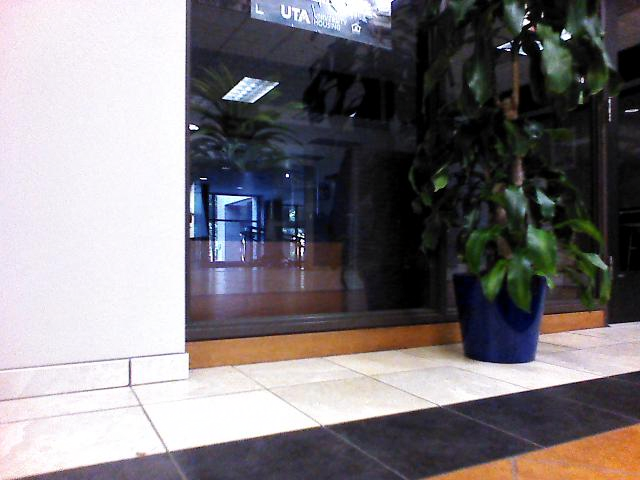}
\end{subfigure}
\hfill
\begin{subfigure}{0.24\linewidth}
\centering
\caption{CASE 3} 
\includegraphics[width=\linewidth]{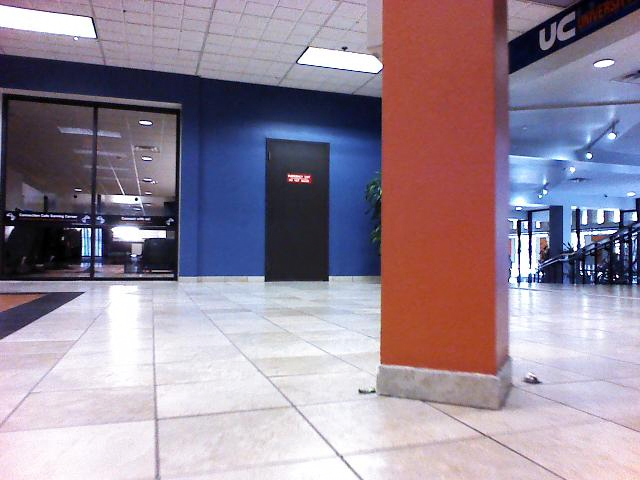}
\end{subfigure}
\hfill
\begin{subfigure}{0.24\linewidth}
\centering
\caption{CASE 4} 
\includegraphics[width=\linewidth]{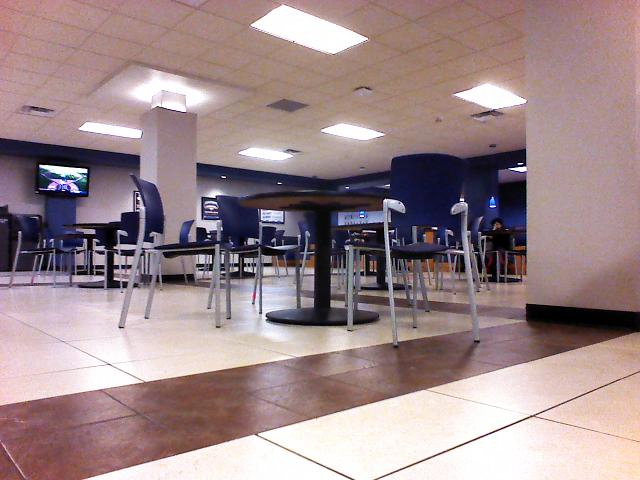}
\end{subfigure}
\caption{Adversarial examples generated by the PGD attack}
\end{subfigure}

\hfill
\begin{subfigure}{\textwidth}  
\centering
\begin{subfigure}{0.24\linewidth}
\centering
 \includegraphics[width=\linewidth]{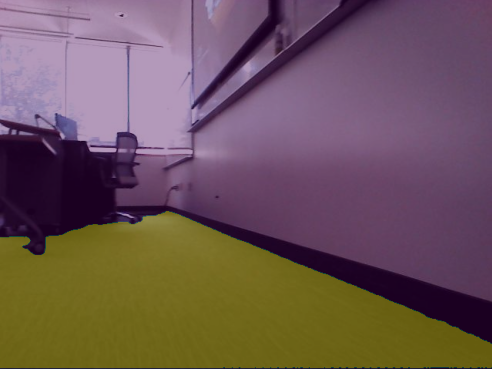}
\end{subfigure}
\hfill
\begin{subfigure}{0.24\linewidth}
\centering
\includegraphics[width=\linewidth]{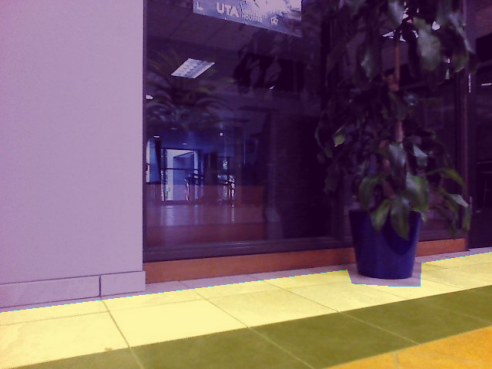}
\end{subfigure}
\hfill
\begin{subfigure}{0.24\linewidth}
\centering
\includegraphics[width=\linewidth]{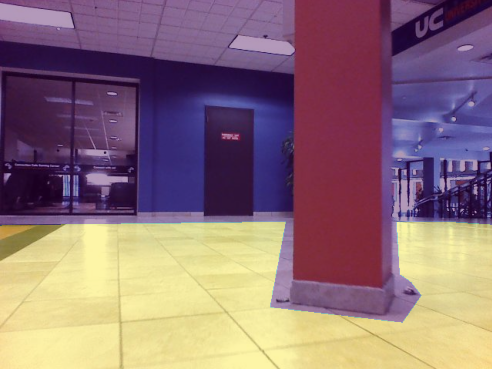}
\end{subfigure}
\hfill
\begin{subfigure}{0.24\linewidth}
\centering
\includegraphics[width=\linewidth]{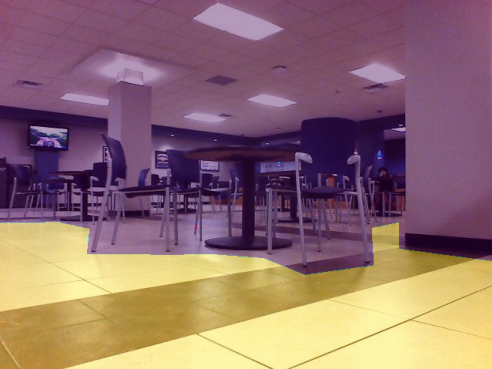}
\end{subfigure}
\caption{Ground truth labels}
\end{subfigure}

\hfill
\begin{subfigure}{\textwidth}  
\centering
\begin{subfigure}{0.24\linewidth}
\centering
 \includegraphics[width=\linewidth]{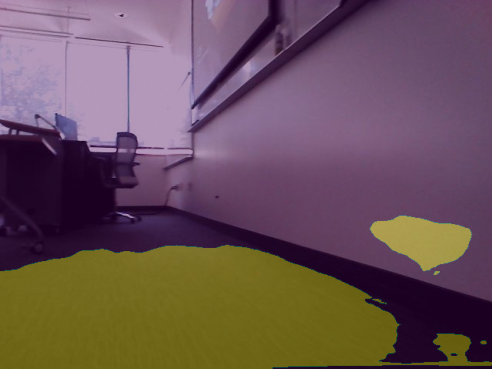}
\end{subfigure}
\hfill
\begin{subfigure}{0.24\linewidth}
\centering
\includegraphics[width=\linewidth]{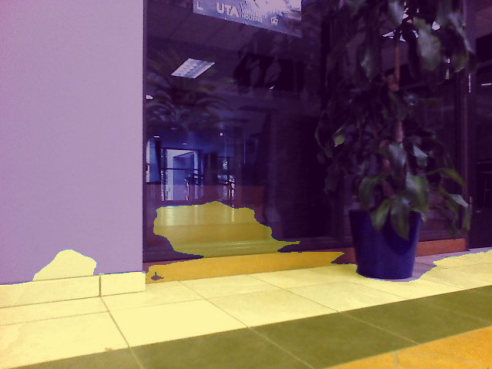}
\end{subfigure}
\hfill
\begin{subfigure}{0.24\linewidth}
\centering
\includegraphics[width=\linewidth]{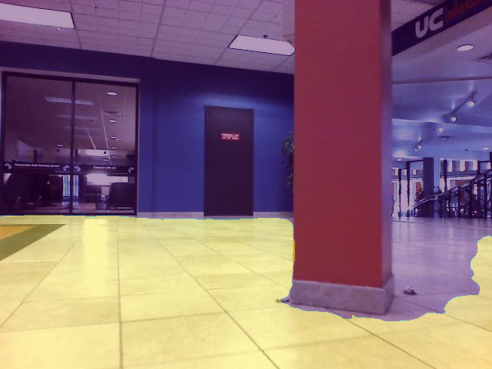}
\end{subfigure}
\hfill
\begin{subfigure}{0.24\linewidth}
\centering
\includegraphics[width=\linewidth]{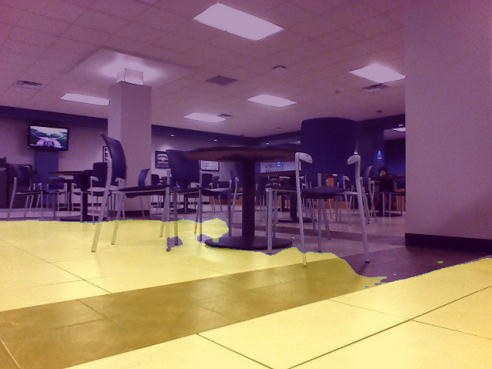}
\end{subfigure}
\caption{Model predictions with no defense}
\end{subfigure}


\hfill
\begin{subfigure}{\textwidth}  
\centering
\begin{subfigure}{0.24\linewidth}
\centering
 \includegraphics[width=\linewidth]{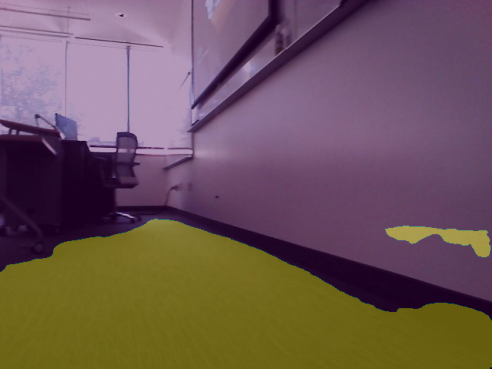}
\end{subfigure}
\hfill
\begin{subfigure}{0.24\linewidth}
\centering
\includegraphics[width=\linewidth]{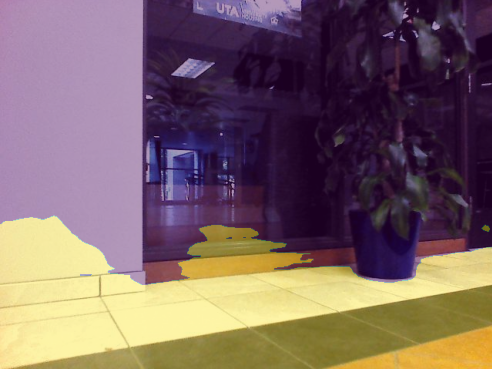}
\end{subfigure}
\hfill
\begin{subfigure}{0.24\linewidth}
\centering
\includegraphics[width=\linewidth]{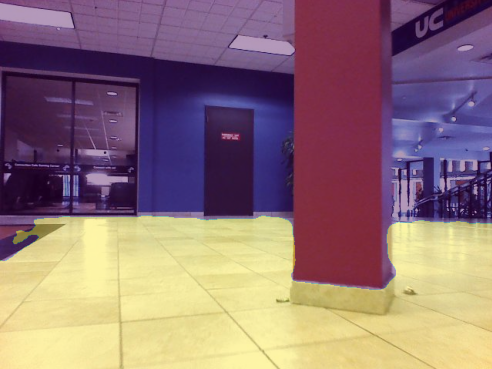}
\end{subfigure}
\hfill
\begin{subfigure}{0.24\linewidth}
\centering
\includegraphics[width=\linewidth]{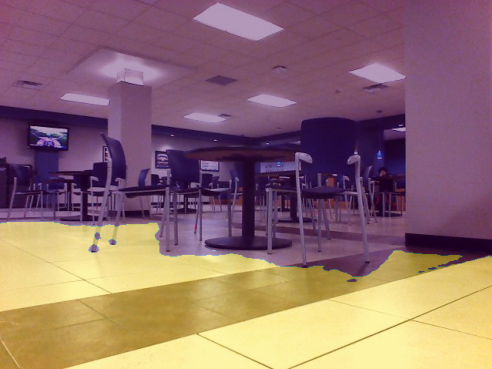}
\end{subfigure}
\caption{Model predictions with traditional adversarial training}
\end{subfigure}


\hfill
\begin{subfigure}{\textwidth}  
\centering
\begin{subfigure}{0.24\linewidth}
\centering
 \includegraphics[width=\linewidth]{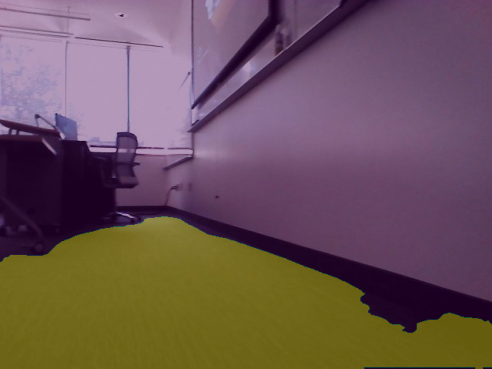}
\end{subfigure}
\hfill
\begin{subfigure}{0.24\linewidth}
\centering
\includegraphics[width=\linewidth]{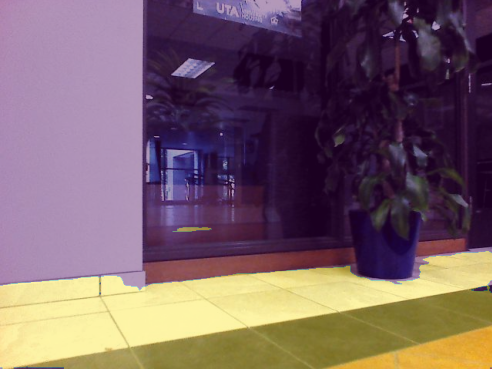}
\end{subfigure}
\hfill
\begin{subfigure}{0.24\linewidth}
\centering
\includegraphics[width=\linewidth]{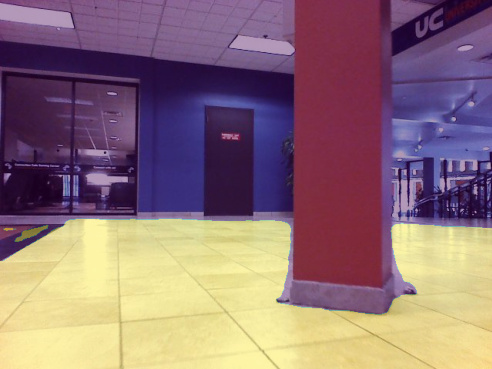}
\end{subfigure}
\hfill
\begin{subfigure}{0.24\linewidth}
\centering
\includegraphics[width=\linewidth]{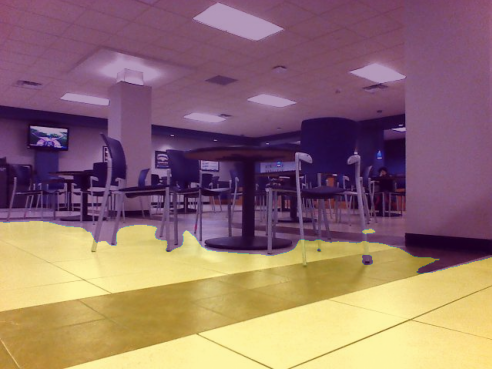}
\end{subfigure}
\caption{Model predictions with adversarial training and loss (our method)}
\end{subfigure}

\caption{Qualitative results}
\label{fig:qual}
\end{figure*}

We train a SegFormer on the positive instances of the dataset and test on the challenging ones. For the experiments, the standard semantic segmentation metrics, Mean Intersection over Union (mIoU) is used.


\subsection{Ablation Study}

An ablation study is conducted to evaluate the performance
of the fine-tuning method on the testing dataset. For the purpose of comparison, we also experiment with two standard semantic-segmentation deep learning baselines, i.e. DeepLabv3~\cite{chen2017deeplab} and PSPNet~\cite{zhao2017pyramid}. Besides, different configurations in terms of employed PGD attack, loss function and Adversarial Training are applied. The findings (Table~\ref{tab:quan}) show that the proposed method, which combines Adversarial Training and the adversarial hidden loss, achieves significant improvements in mIoU scores, indicating its potential to enhance the model's robustness. These findings are in consensus with the hypotheses that 1) During adversarial training the model learns to generalize better and becomes less sensitive to small perturbations in input data and 2) by incorporating losses in the hidden layers, the model gains the capability to resist perturbations not only at the input but also within intermediate representations, rendering it more resilient.

\subsection{Qualitative results}
Figure~\ref{fig:qual} illustrates qualitative results obtained from the test dataset, which depicts challenging instances. We compare the performance of our approach, which combines Adversarial Training (AT) with adversarial hidden loss ($l_h$) to enhance resistance against perturbations (PGD), and an approach solely employing Adversarial Training. The PGD attack can either expand the free-space into nearby walls and obstacles or transform portions of free-space into obstacles. Moreover, it is observed that AT partially restores the areas that were transformed into obstacles by the PGD attack. Overall, the combined approach significantly enhances robustness by effectively correcting the introduced false free-space, as demonstrated in the results, compared to using Adversarial Training alone. An erroneous case  that the proposed method fails is also presented (Case 4). This is owing to the presence of certain objects (chairs with thin legs), which creates visual similarities with free space regions, consequently leading to errors in prediction. Furthermore, another reason is the poor performance of the clean model (model with no defense) that fails to identify free-space due to the presence of clutter objects.

 \subsection{Effect of regularization strength}

Figure~\ref{fig:lambda} presents the SegFormer's performance for different $\lambda$ values. It is observed that when $\lambda$ is no more than 1, regularization can enhance the model's robustness, whereas for $\lambda$ = 10, the performance drops dramatically. This phenomenon is attributed to excessive regularization, which can lead to the over-smoothing of hidden layers.

\begin{figure}[h]
\centering
\includegraphics[width=0.8\linewidth]{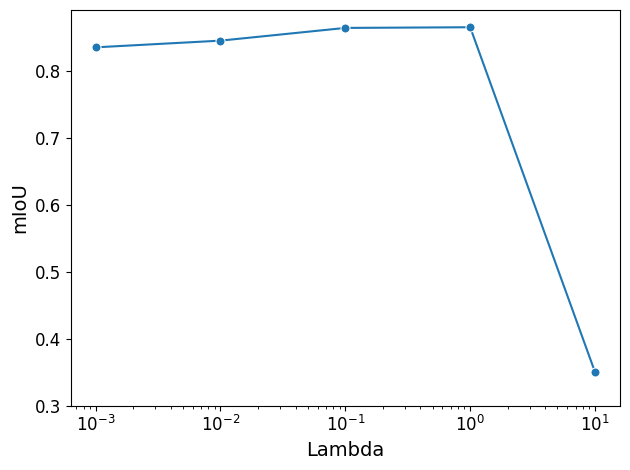}
\caption{mIoU vs regularization intensity $\lambda$ }
\label{fig:lambda} 
\end{figure}

\section{Conclusions and Future Work}
This study aims to enhance the robustness of free-space segmentation against adversarial perturbations. We introduce a novel method that combines Adversarial Training (AT) with an adversarial hidden loss to reinforce model resistance to perturbations generated by the Projective Gradient Descent (PGD) attack. Incorporating adversarial examples into the training process, exhibits the ability to improve the model's generalization and mitigates sensitivity to minor input perturbations. Results demonstrate a notable improvement when using the combined approach compared to solely implementing the traditional AT. Nevertheless, challenges remain in cases where objects create visual similarities with free-space regions, leading to prediction errors. Future research directions include incorporating object detection modules in our approach as well as experimenting with various novel adversarial defense strategies.


\bibliographystyle{IEEEtran}
\bibliography{main}

\end{document}